\documentclass[conference]{IEEEtran}
\IEEEoverridecommandlockouts
\usepackage{cite}
\usepackage{amsmath,amssymb,amsfonts}
\usepackage{algorithmic}
\usepackage{graphicx}
\usepackage{textcomp}
\usepackage{xcolor}
\usepackage{multicol}
\usepackage{caption}
\usepackage{xfrac}
\usepackage{siunitx}
\usepackage[utf8]{inputenc}
\usepackage{caption,subcaption}
\def\BibTeX{{\rm B\kern-.05em{\sc i\kern-.025em b}\kern-.08em
    T\kern-.1667em\lower.7ex\hbox{E}\kern-.125emX}}
\usepackage{booktabs} 
\usepackage{multirow}
\begin{document}

\title{Machine Learning based System for Vessel Turnaround Time Prediction

\thanks{
\textcopyright 2020 IEEE. Personal use of this material is permitted. Permission from IEEE must be obtained for all other uses, in any current or future media, including reprinting/republishing this material for advertising or promotional purposes, creating new collective works, for resale or redistribution to servers or lists, or reuse of any copyrighted component of this work in other works.
}
}

\author{\IEEEauthorblockN{Dejan Štepec\IEEEauthorrefmark{1},
Tomaž Martinčič\IEEEauthorrefmark{1}, Fabrice Klein\IEEEauthorrefmark{2}, Daniel Vladušič\IEEEauthorrefmark{1} and
Joao Pita Costa\IEEEauthorrefmark{1}}
\IEEEauthorblockA{\IEEEauthorrefmark{1}XLAB Research,
Ljubljana, Slovenia\\}
\IEEEauthorblockA{\IEEEauthorrefmark{2}Port of Bordeaux,
Bordeaux, France\\}
}


\maketitle

\begin{abstract}
In this paper we present a novel system for predicting vessel turnaround time, based on machine learning and standardized port call data. We also investigate the use of specific external maritime big data, to enhance the accuracy of the available data and improve the performance of the developed system. An extensive evaluation is performed in Port of Bordeaux, where we report the results on 11 years of historical port call data and provide verification on live, operational data from the port. The proposed automated data-driven turnaround time prediction system is able to perform with increased accuracy, in comparison with current manual expert-based system in Port of Bordeaux.
\end{abstract}


\section{Introduction}

 The maritime industry has seen a considerable growth in recent years across all the major cargo types, with more than 90\% of the world’s trade carried by sea~\cite{imo}.
 In fact, according to the latest United Nations Conference on Trade and Development (UNCTAD)~\cite{unctad_review}, maritime trade is expected to expand at an average annual growth rate of 3.5\% between 2019 and 2024.  According to the UNCTAD report~\cite{unctad_review}, there were over 95.000 ships in early 2019 and there is a stable increase of the total capacity over the last decades. The increasing number of ships and their capacity is putting pressure on ports, especially on port turnaround times, which is according to the UNCTAD~\cite{unctad_review}, one of the main indicators of ports efficiency and trade competitiveness. UNCTAD now performs novel analysis using Automatic Identification System data (AIS)~\cite{port_time}, to calculate the vessel turnaround time, based on vessel type and country. According to the UNCTAD report~\cite{unctad_review}, ships spent in port a median time of 23.5 hours (0.97 days), during one port call. Dry bulk carriers spent the most time in port, with a median of 2.05 days, almost three times the median time of a container ship. Reducing the time in port, reflects in the ability to accommodate more ships, resulting in an increased efficiency of the whole maritime chain.

Ports internal data can provide plenty of information to analyze and optimize operations in the port. One of the richest and most general information, standardized across all of the ports, is that of port calls. Vessels are usually announced at least 24 hours prior to the port arrival. This is not only due to port authority regulations, but also due to national maritime agencies, that require this kind of information to be submitted before arrival to their territorial waters. The forms that need to be submitted are standardized and were agreed in a Convention on Facilitation of Maritime Traffic (FAL)~\cite{fal_forms}. One of the main objectives was to standardize the procedures and declarations that need to be submitted to the public authorities in order to prevent unnecessary delays in maritime traffic. Additionally, business metrics and data driven decisions can be made, based on the extracted insights from the port call data. Detailed descriptions of the loaded and unloaded cargo and the amount of it, together with arrival and departure times, can be directly used to provide port operators with more accurate predictions of turnaround times.

In this work we explore the use of FAL forms data, captured in Port Community Systems (PCS) for machine learning (ML) based vessel turnaround time prediction. PCSs are robust, proprietary legacy port IT systems~\cite{pcs}, that are connected to governmental, as well as other maritime business entities, in order to reduce clerical burden and encourage multiple use of data. The problem tackled in this work, can also be described as predicting vessel estimated time of departure (ETD), as we are able to provide this information directly from the predicted vessel turnaround time. FAL forms data represents standardized data, captured similarly in all of the ports around the world. Rich information about the port call, including the type and amount of loaded and unloaded cargo represents an untapped potential not only for the case of ETD prediction, but also for a wider port operations analysis, which could provide actionable insights and is also demonstrated in our work. We also investigate the use of external environmental data, such as weather data, as well as Port of Bordeaux specific tidal level data. Additionally, we explore the use of AIS data to facilitate real-time retrieval of vessel entry and exit times, as well as to mitigate the problems of missing data. Our main contributions can be summarized as follows:
\begin{itemize}  
    \item To the best of our knowledge, we present the first ML based system for vessel turnaround time prediction, based on FAL forms operational data.
    
    \item We perform an extensive exploratory data analysis (EDA) and feature engineering, offering a level of explainability, by measuring feature importance.
    
    \item We evaluate state-of-the-art ML methods on large-scale historical, as well as operational (live) data from the Port of Bordeaux.
    
    \item We investigate the use of external data, such as weather, tidal and AIS data, in order to improve the performance and robustness of the proposed system.
\end{itemize}

\section{Related Work}

Optimizing vessel turnaround time is a well-known research problem, that has been tackled implicitly in the literature and in practice by optimizing berth scheduling~\cite{berth1, berth2, berth3, berth4}. Berth scheduling is usually defined as a multi-objective optimization problem, by optimizing different port business performance indicators (e.g. vessel turnaround time, revenue, customer satisfaction, utilization of the equipment). Some of the objectives are conflicting and optimal decisions need to be taken - the so called Pareto optimal. Such modeling tasks are usually performed with certain assumptions. The three most critical assumptions, according to~\cite{berth3} are related to berthing space (continuous vs. discrete), vessel arrivals (static vs. dynamic) and handling time (static vs. dynamic). Berths are usually modeled in a discrete space, as a finite set. In a static vessel arrival modelling, all the vessels are assumed to be already in the port, while with dynamic modelling, not all the vessels have arrived, but their arrival times are known in advance. Static handling time assumes that handling time will be provided manually, while in a dynamic setting, the handling time is modelled as a function of some operational parameters (e.g. number of quay cranes). In comparison with berth scheduling modeling objectives, we address vessel turnaround time directly, based on standardized operational live and historical data from the ports.

The proposed vessel turnaround time prediction system presents an independent, pluggable unit, that can be used as a standalone solution, as part of the existing Port Community Systems (PCS), or connected and also available to other maritime transport stakeholders. Unreliable arrival and handling times of a vessel have also been recognized in the berth scheduling research~\cite{berth5}, where time buffers were considered to be inserted between the vessels, occupying the same berth. In comparison, the proposed system can be used in existing berth scheduling solutions, by replacing simple sub-optimal functions for handling time modeling (e.g. number of quay cranes) or time buffering solutions, with the actual operational, data driven solutions for vessel turnaround time prediction. Proposed system will also complement already available systems for vessel arrival prediction (ETA), also based on FAL forms data~\cite{eta1_fal}.

Automatic Identification System (AIS) data from the vessels is also being utilized in the maritime domain for maritime traffic pattern analysis~\cite{ais_pattern}, as well as in real-time based systems for vessel destination and ETA prediction~\cite{debs_ais, ais_eta}. We also make use of AIS data in our proposed ETD prediction system, to supplement missing data about vessel entry or exit times. Real-time AIS data also improves the timeliness of ETD predictions. Similar to works on ETA~\cite{eta_rotterdam, eta_long}, we also utilize external meteorological and Port of Bordeaux specific tidal level data.


\section{Port of Bordeaux}
\label{sec:GPMB}

The Port of Bordeaux (GPMB) is located on the Atlantic coast, just outside Bordeaux, whose population will shortly reach 1 million. GPMB is the focal point of a dense river, sea, air, rail and road traffic network. GPMB is a core port of the Trans-European Network of Transport (TEN-T) and belongs to the Atlantic Corridor. 
GPMB ranks 7th of French ports, as it totals 2\% of French maritime traffic (i.e. 8 to 9 Mt/year), heavily based on hydrocarbon goods (fuels) and cereals. Severe environmental regulations lead to a demanding operational processes, allied with a strong financial pressure for the port authority. 
GPMB has developed its own Port Community System (VIGIEsip) in 2014 to organize port calls, to bring digital services to the port and to comply with the single window directive. This important step of the port digitalization implements the PCS approach as in \cite{pcs}, taking into account the specific challenges of the port and is also the adopted PCS by 14 other French ports.

Port of call means an intermediate stop for a ship on its scheduled journey. It means also hours of idle time for ships during the processing of the cargo. Vessel idle time has a significant cost for the customers of the port, as it reduces the vessel utilization and cargo turnover rate. In addition, there may be unforeseen circumstances, delaying the departure of the ship. This is the reason why the ports seek better accuracy on this knowledge, aiming to reduce the turnaround time.
The turnaround time is one of the least well-known information sources in GPMB. It is difficult to properly determine this measure because GPMB is in an estuary and the internal and external information it relies on (e.g. commodity category, amount of cargo, tidal levels and weather), is hard to reliably incorporate into current simple manual models. The current approach relies on static GPMB performance metrics for specific cargo processing capabilities (e.g. tonnage per hour for a specific cargo, available manpower), cargo tonnage to be processed and the next high tide time. Shipping agents provide this information through VIGIEsip, after discussion with terminal operators, which provide information about the available manpower and equipment, that can be allocated to the port call.

Uniqueness of the GPMB, their specific location and non-deterministic port call operations (first come, first served model) represents a significant challenge, indicating an upper bound for the predictive performance of turnaround time prediction. In comparison with currently adapted manual models, we propose to utilize a vast amount of historical port call data, captured in VIGIEsip, to increase GPMB operational efficiency. The use of standardized data ensures the applicability of the proposed solution also to all the other ports, with the same proposed methodology.

\section{Methodology}

\begin{figure*}[htbp]
    \centering
      \includegraphics[width=\linewidth]{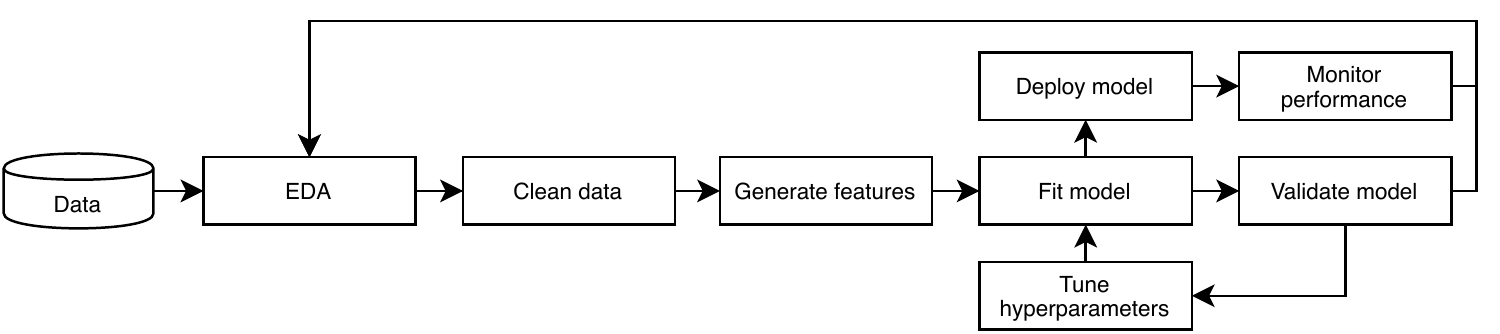}
      \caption{Developing a predictive model on structured data consists out of several steps. Data preparation, which includes data cleaning, exploratory data analysis and feature engineering, represents the majority of the machine learning workflow. This then presents an input to the machine learning model, which trained model is evaluated on left-out validation and test data. The process is usually repeated in an incremental way, to improve the performance of the deployed model.}
      \label{fig:GPMB_ML_pipeline-high_level}
\end{figure*}

\subsection{Input Data}

Port call data, as captured in the PCS of the Port of Bordeaux, VIGIEsip, represents the main source of the data. We collected 11 years of historical port call data, between the beginning of 2008 and the end of 2018. The data represents a subset of the data, that is extracted from the FAL forms (FAL form 1 and 2) and the main attributes represent the vessel unique identifiers, amount and the type of the loaded and unloaded cargo, arrival berths and the exact times of arrival and departure. In total, we collected 6055 port calls, which consisted out of 1905 unique vessels, unloading 55 and loading 46 types of unique cargo types. Average turnaround time of the port calls in historical data is 53 hours. Additionally, GPMB provided live access to port call data, including port turnaround time predictions, which was used to evaluate the proposed method on live operational data. In the live operational data we noticed missing entry and exit times, or their delayed availability. To solve the problem, we combined port call data with AIS data, obtained from AISHub\footnote{http://www.aishub.net/}. Obtaining vessel arrival and departure times from AIS data represents an universal approach, providing real-time availability of arrival and departure data and also serves as a verification of FAL forms data.

\begin{figure}[ht!]
\centerline{\includegraphics[width=0.5\textwidth]{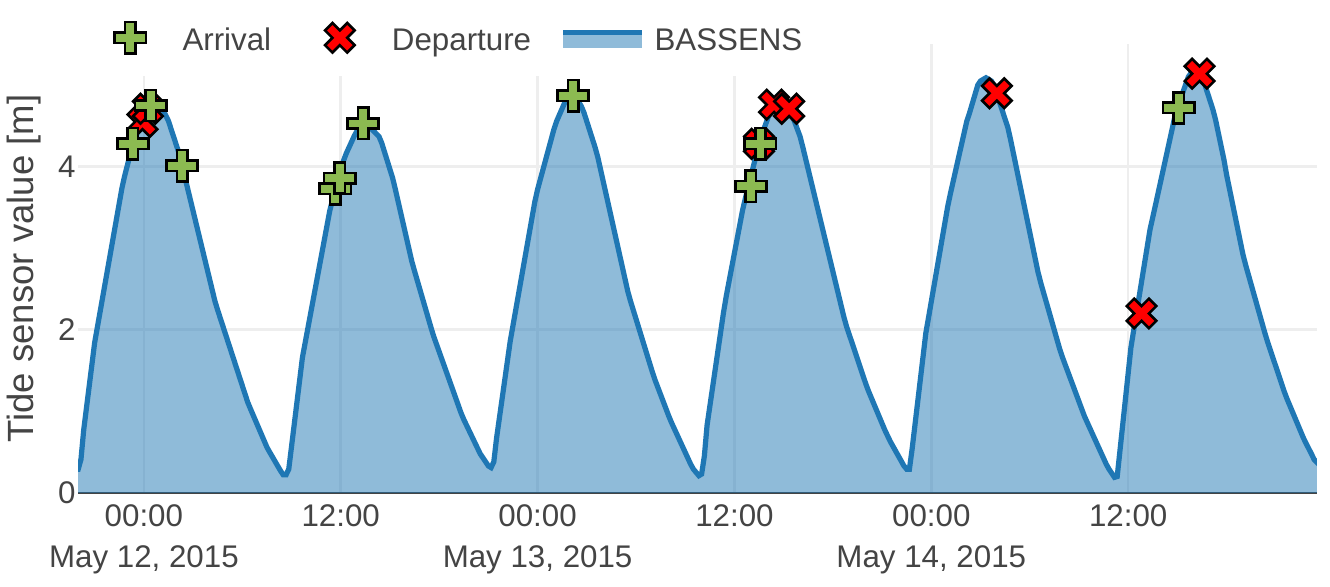}} 
\caption{Vessels arrivals, departures and water height in Bassens for 4 days. Tidal level has a significant impact on turnaround time, as the vessels arrive and depart predominately on high tides.}
\label{fig:gpmb_tides_arrivals_departures}
\end{figure}

We also combine the available maritime data with external weather and tidal information data. Our hypothesis is that weather has influence on port operations: especially strong winds affecting crane operations and rain affecting dry cargo operations. We obtained hourly aggregated historical weather data (temperature, wind, precipitation) from the Dark Sky\footnote{https://darksky.net/}. Due to the specifics of GPMB (located about 90 km deep in the Gironde estuary) and their dependence on tidal levels - presented in Figure~\ref{fig:gpmb_tides_arrivals_departures}, we also include tidal level information from multiple sensors along the Gironde estuary.
    
Based on a list of holidays in France\footnote{https://github.com/dr-prodigy/python-holidays}, for all the years of the historical data, we evaluated their influence on turnaround times. Because of the average turnaround time of more than 2 days, we added Boolean features, indicating the presence of a holiday in \textpm 3 days from the vessel arrival.

\subsection{Data Preparation}
\label{sec:data_preparation}

\begin{figure*}[htbp]
    \centering
    \begin{subfigure}{0.49\textwidth}
      \includegraphics[width=\linewidth]{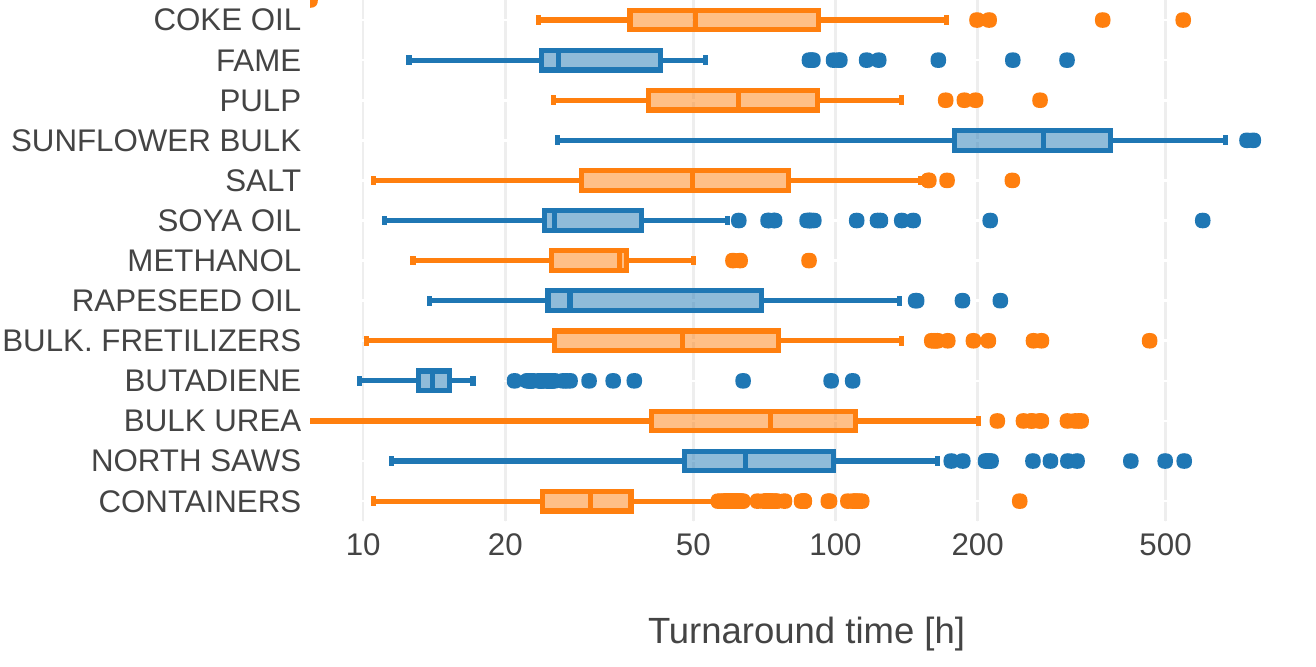}
      \caption{}
      \label{fig:gpmb_unloading_cargo_types_hours_stay}
    \end{subfigure}\hfil
        \begin{subfigure}{0.49\textwidth}
      \includegraphics[width=\linewidth]{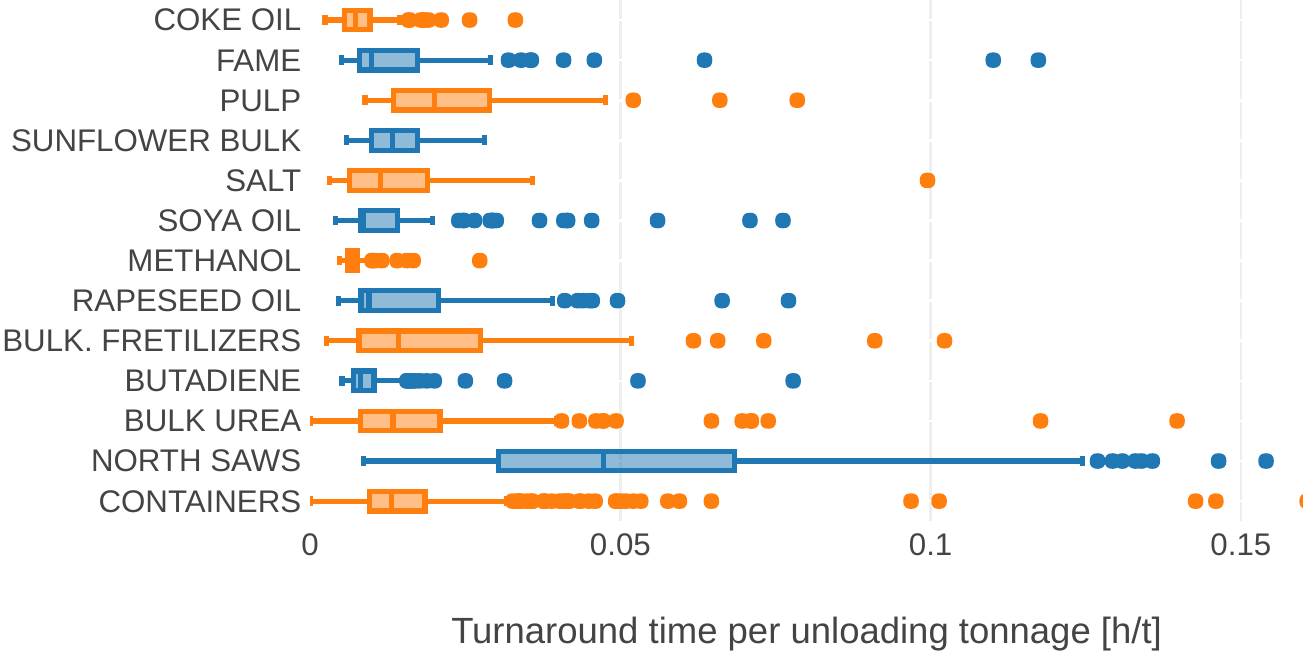}
      \caption{}
      \label{fig:gpmb_unloading_cargo_types_hours_stay_normalised}
    \end{subfigure}\hfil
    \caption{Influence of the most frequent unloading cargo types on the turnaround time. The visualization represents variability of the amount of cargo on the vessel (a) and per-tonnage normalized processing time (b).}
    \label{fig:vae_anomaly}
\end{figure*}

The data that was used, represents operational data from GPMB, thus some data preparation is needed for stable model training and increased generalization performance. We first removed all the vessels without any cargo to load and unload (i.e. empty ships). We also removed ships, whose turnaround time was less than 1 hour and those of which turnaround time exceeded two times the standard deviation from the median, of the corresponding cargo type (i.e. outliers, corresponding to erroneous inputs in VIGIEsip). We also removed vessels calls, with combinations of loading and unloading cargo types, that occurred in less than 5 occurrences. This filtering procedure reduced the number of port calls to 5492, with 1768 unique vessels and 34 unloading and 32 loading cargo types, respectively.

Data cleaning and constructing meaningful features, that hold predictive value, represents a major part of the machine-learning workflow, presented also in Figure~\ref{fig:GPMB_ML_pipeline-high_level}. We investigate the data in exploratory data analysis, where the influence of different parameters on turnaround time is analysed. Some results of this analysis are presented in section~\ref{subsection:EDA}. Besides identifying predictive features, encoding them appropriately is also very important. Timestamp data (e.g. arrival time) needs to be transformed into multiple categorical features (e.g. day of the week, hour). Most of the machine learning models also require the data to be normalized and categorical features transformed into numerical ones (e.g. one-hot-encoding). We instead used the method, described in the next section, that omits most of such requirements and provides out-of-the-box state-of-the-art performance.

\subsection{Predictive Model}

We utilized recently presented gradient boosting based method CatBoost~\cite{catboost}, which properties offer the most convenient use in operational environments with heterogeneous, structured data. The CatBoost method uses decision trees as base predictors and thus omits the need for data normalization as the preprocessing step. CatBoost also handles categorical features during training, as opposed to preprocessing time, thus omitting the need for special transformations. Decision trees also have the increased level of explainability of the results, in comparison with other black-box methods (e.g. neural networks). CatBoost also outperforms other state-of-the-art gradient boosted decision methods (e.g. XGBoost~\cite{xgboost}) in terms of predictive accuracy and speed, out-of-the-box, without the need for extensive hyperparameter fine-tuning.\footnote{https://catboost.ai} Nevertheless, we analyse the importance of hyperparameters on our data, using Grid Search parameter tuning procedure, provided with the open-source CatBoost implementation\footnote{https://github.com/catboost/catboost} and present the results in section~\ref{results}.

\subsection{Evaluation}
\label{sec:evaluation}

To evaluate the proposed method, we used a combination of historical and live operational data from the port. Turnaround time is computed as a difference between arrival and departure times. To effectively use the available data, we evaluated the method using the cross-validation procedure with a 1 year left-out strategy on historical data. In this way, the proposed method was evaluated on all of the 11 years of data. We evaluated the method on the splits, using the Mean Absolute Error (MAE), Root Mean Square Error (RMSE) and Mean Absolute Percentage Error (MAPE). With such combination of evaluation metrics, we ensure explainability of the performance (MAE, MAPE), as well as to evaluate the influence of the large errors (RMSE). The best performing model on historical data was also evaluated on live operational data of 93 vessels from the port. In this case, the comparison was also made against the expertly provided predictions from the port, based on their current manual approach, described in section~\ref{sec:GPMB}.

\section{Results}
\label{results}

\subsection{Exploratory Data Analysis}
\label{subsection:EDA}

\begin{table*}
           \centering
           \captionsetup[subtable]{position = below}
          \captionsetup[table]{position=top}
           \caption{Evaluation on historical data for 10 most frequent unloading (U) cargo types (a) and loading (L) cargo types (b).}
           \begin{subtable}{0.45\linewidth}
               \scriptsize
\centering
\caption{}
\begin{tabular}{l|rr|r|r}
\toprule
                     & \multicolumn{2}{c|}{\fontsize{7}{8}\selectfont MAE [h]} &       \multicolumn{1}{c|}{\fontsize{7}{8}\selectfont RMSE} &     \multicolumn{1}{c}{\fontsize{7}{8}\selectfont MAPE [\%]} \\
Unloading cargo type & \fontsize{7}{8}\selectfont CatBoost & \fontsize{7}{8}\selectfont Linear R. & \fontsize{7}{8}\selectfont CatBoost & \fontsize{7}{8}\selectfont CatBoost \\
\midrule
           BUTADIENE &     \textbf{2.51} &             4.46 &     4.33 &    13.87 \\
            METHANOL &     \textbf{2.69} &             6.18 &     4.25 &     8.51 \\
          SOYA OIL &     \textbf{7.16} &             8.47 &    13.28 &    22.49 \\
          CONTAINERS &     \textbf{7.23} &             8.42 &     9.79 &    22.81 \\
         RAPESEED OIL &    \textbf{11.12} &            11.68 &    17.84 &    22.81 \\
                 SALT &    \textbf{16.67} &            19.30 &    25.40 &    41.15 \\
     BULK. FRETILIZERS &    \textbf{17.49} &            18.76 &    27.87 &    34.84 \\
           BULK UREA &    \textbf{23.19} &            26.70 &    32.03 &    37.07 \\
  NORTH SAWS &    \textbf{24.87} &            26.83 &    33.31 &    45.84 \\
      SUNFLOWER BULK &    83.03 &            \textbf{82.37} &   105.47 &    41.53 \\
      
\midrule
Top 10 cargo types (U)  &  \textbf{14.62} & 16.33 & 27.46 & 28.43 \\
All cargo types (U)             &   \textbf{15.75} &   17.55 &   28.22 &   30.46 \\

\bottomrule
\end{tabular}
\label{tbl:unloading_cargo_type_mae}
           \end{subtable}%
           \hspace*{4em}
           \begin{subtable}{0.45\linewidth}
               \scriptsize
\centering
\caption{}
\begin{tabular}{l|rr|r|r}
\toprule
                       & \multicolumn{2}{c|}{\fontsize{7}{8}\selectfont MAE [h]} &      \multicolumn{1}{c|}{\fontsize{7}{8}\selectfont RMSE} &     \multicolumn{1}{c}{\fontsize{7}{8}\selectfont MAPE [\%]} \\
    Loading cargo type & \fontsize{7}{8}\selectfont CatBoost & \fontsize{7}{8}\selectfont Linear R. & \fontsize{7}{8}\selectfont CatBoost & \fontsize{7}{8}\selectfont CatBoost \\
\midrule
            CONTAINERS &     \textbf{7.24} &             8.41 &     9.79 &    22.79 \\
             BULK CORN &     \textbf{8.65} &            10.51 &    13.89 &    29.64 \\
 PROD.CHIM.LIQ. AUTRES &     \textbf{9.42} &            11.48 &    15.58 &    32.78 \\
              BULK WHEAT &    \textbf{10.04} &            11.53 &    14.98 &    31.34 \\
       OTHER MINERALS &    \textbf{12.12} &            13.55 &    17.53 &    38.25 \\
        SUNFLOWER BULK &    \textbf{13.34} &            14.62 &    21.41 &    38.88 \\
       SUNFLOWER OIL &    13.64 &            \textbf{13.54} &    18.58 &    29.27 \\
       SCRAP &    \textbf{22.60} &            26.50 &    32.22 &    34.15 \\
        SUNFLOWER PELLETS &    \textbf{23.06} &            24.47 &    33.02 &    35.75 \\
   FAME  &    \textbf{23.61} &            28.83 &    39.77 &    42.34 \\
   
\midrule
Top 10 cargo types (L)  &   \textbf{10.64} &   12.37 &   17.93 &   29.53 \\
All cargo types (L)             &   \textbf{11.57} &   12.74 &   19.99 &   30.81 \\
\bottomrule
\end{tabular}
\label{tbl:loading_cargo_type_mae}
           \end{subtable}
           \label{results_historical}
       \end{table*}
       
\begin{figure}[ht!]
\centerline{\includegraphics[width=0.5\textwidth]{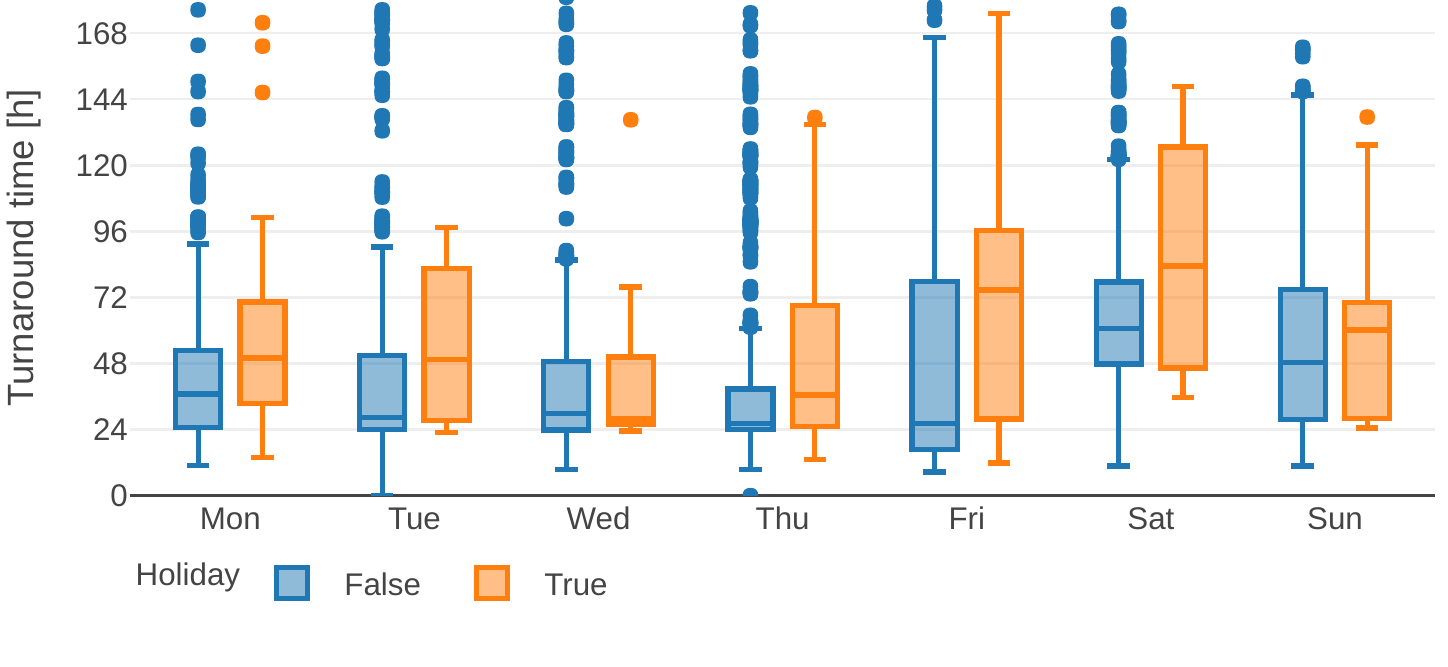}} 
\caption{The influence of the arrival day and holidays on the turnaround time. The increase is visible at the end of the week and on holidays.}
\label{fig:gpmb_hours_stay_per_arrival_day}
\end{figure}

In order to develop accurate predictive models for a specific domain, understanding of the target domain is needed, as well as the influence of operational parameters on the target variable to be predicted (i.e. turnaround time). This is achieved by the know-how of domain experts and the analysis of the available historical data. We first analyse how specific cargo to be loaded or unloaded influences turnaround time. As presented in Figure~\ref{fig:gpmb_unloading_cargo_types_hours_stay}, different cargo types have different processing times, variability for a specific cargo type also varies. This is partially due to the variability of the amount of the cargo. We also report the same results, with per tonnage normalized values, presented in Figure~\ref{fig:gpmb_unloading_cargo_types_hours_stay_normalised}. We can see that there are certain cargo types, that have standardized procedures, with less variance (e.g. containers, liquid cargoes, gas) and some, especially bulk cargo types, which per tonnage processing time shows greater variability. Such findings indicate, that predictive performance will differ along different cargo types.



The day of the arrival to the port also has a significant impact on the turnaround time, as presented in Figure~\ref{fig:gpmb_hours_stay_per_arrival_day}. Turnaround time increases on Friday and Saturday, due to limited port operational capacity. Some port operators are also not operating during the weekend, due to increased labour cost. Some ship operators also do not want to pay higher fees during the weekend. There are also contractual obligations for certain ship operators and cargoes, which have priority, due to emergency, or just-in-time deliveries for the factories. Holidays and consequently reduced manpower, or the event of a closed port, also increases turnaround times.

\subsection{Predictive Modelling}


We first evaluate predictive performance on large-scale historical data with a cross-validation procedure and evaluation metrics as described in section~\ref{sec:evaluation}. We report the results in Tables~\ref{tbl:unloading_cargo_type_mae} and~\ref{tbl:loading_cargo_type_mae} for the most frequent unloading and loading cargo types, respectively. We also compare our results with the linear regression model, to demonstrate superiority of the used CatBoost method~\cite{catboost}, over baseline machine learning models. Note that careful feature normalization and categorical feature transformation was performed for linear regression, as described in section~\ref{sec:data_preparation}. We can see that overall MAE is around 15 hours for unloading and 11 hours for loading operations, or expressed with MAPE - around 30\% error. The comparison is made against ground truth turnaround time, computed out of exact arrival and departure times. The error on predicted turnaround time reduces greatly for certain cargo types (e.g. liquid cargo, containers), even below 10\% and it is evident, that the error correlates with specific cargo types and findings presented in EDA section~\ref{subsection:EDA}. Overall error is similar for both, unloading and loading operations and CatBoost~\cite{catboost} significantly outperforms baseline linear regression method.

We additionally evaluate the deployed system on 2 months worth of live operational data. The data consisted of 93 port calls, with at least 3 arrivals for each cargo type, in order to have reliable statistics, comparable with historical data. We have used best performing CatBoost model from offline evaluation on historical data. Results, presented in Table~\ref{tbl:live_results} show, that the performance (MAE) is consistent with the performance on offline data and consistently better from the current simplistic model used by the port, by a large margin.

\begin{table}
\caption{Evaluation on live operational data for cargo types with at least 3 arrivals in a evaluation period of 2 months. Results are reported for unloading and loading operation, as well as overall results for separate operations.}
\label{tbl:live_results}
\scriptsize
\centering
\begin{tabular}{llrr}
\toprule
        &   Cargo type  &  CatBoost MAE [h] &  Port MAE [h] \\
\midrule
\multirow{10}{*}{\rotatebox[origin=c]{90}{Unloading}} & ENGR.LIQUIDES &           \textbf{2.03} &           18.66 \\
        & METHANOL &           \textbf{2.34} &            9.49 \\
        & RAPESEED OIL &           \textbf{2.58} &           32.16 \\
        & BUTADIENE &           \textbf{3.39} &           16.73 \\
        & SOYA OIL &           \textbf{6.84} &           24.74 \\
        & TALL OIL &           \textbf{7.32} &           25.49 \\
        & BULK UREA &          \textbf{16.93} &           57.29 \\
        & NORTH SAWS &          \textbf{25.37} &           55.48 \\
        & SUNFLOWER BULK &          \textbf{89.77} &          188.93 \\
\cmidrule{2-4}
 & Combined & \textbf{16.52} & 45.81 \\
\midrule
\multirow{6}{*}{\rotatebox[origin=c]{90}{Loading}} & MAIS VRAC &           \textbf{8.35} &           19.41 \\
        & CRUSHED TYRES &          10.75 &            \textbf{9.75} \\
        & BULK WHEAT &          \textbf{10.78} &           17.72 \\
        & SUNFLOWER OIL &          \textbf{12.68} &           14.00 \\
        & SCRAP &          \textbf{13.56} &           56.88 \\
\cmidrule{2-4}
 & Combined & \textbf{9.97} & 22.75 \\
\bottomrule
\end{tabular}
\end{table}




\subsection{Ablation Study}

    \begin{table}
    \caption{Features used in the best performing model and their importance. Feature importance values are normalized so that the sum of importances of all features is equal to 100.}
    \label{tbl:features_importance}
    \centering
    \begin{tabular}{lr}
    \toprule
                         Feature &  Importance \\
    \midrule
            cargo type (U) &       17.40 \\
               cargo tonnage (U) &       16.15 \\
                  day of entry &       12.71 \\
                 berth (U) &       11.13 \\
              cargo type (L) &        8.54 \\
     hour of entry (round 4) &        8.21 \\
                  berth (L) &        8.03 \\
       fiscal cargo type (L) &        6.70 \\
     fiscal cargo type (U) &        5.56 \\
                 cargo tonnage (L) &        4.72 \\
               holiday on entry &        0.31 \\
                holiday in 2 days &        0.20 \\
               holiday 1 day ago &        0.18 \\
                holiday in 1 day &        0.16 \\
    \bottomrule
    \end{tabular}
    \end{table}

The predictive models, presented in Tables~\ref{results_historical} and~\ref{tbl:live_results} were only using basic FAL forms vessel and cargo specific data (i.e. unloading and loading cargo types, amount of cargo, arrival temporal information and berthing information) and holiday information. The used features and their importance\footnote{https://catboost.ai/docs/concepts/fstr.html} are presented in the Table~\ref{tbl:features_importance}. We also experimented with other features, that naturally should have the influence on turnaround time, but it did not improve the predictive performance of our model. Tidal levels, as presented in Figure~\ref{fig:gpmb_tides_arrivals_departures} have the influence on arrivals, but encoding water height and the time since the last high and low water, did not improve the results of our model. Similarly, the congestion in the port should have an influence on the turnaround time. We experimented with multiple features, that encoded congestion (e.g. number of the vessels in the port, number of vessels with the same (U/L) cargo, average turnaround time for the last N ships that visited the port in the last M days), but did not see the improvement in the predictive performance. Note that these conclusions relate to GPMB data, which is a relatively small port. We argue that such features should be useful in general.

\begin{figure}[ht!]
\centerline{\includegraphics[width=0.5\textwidth]{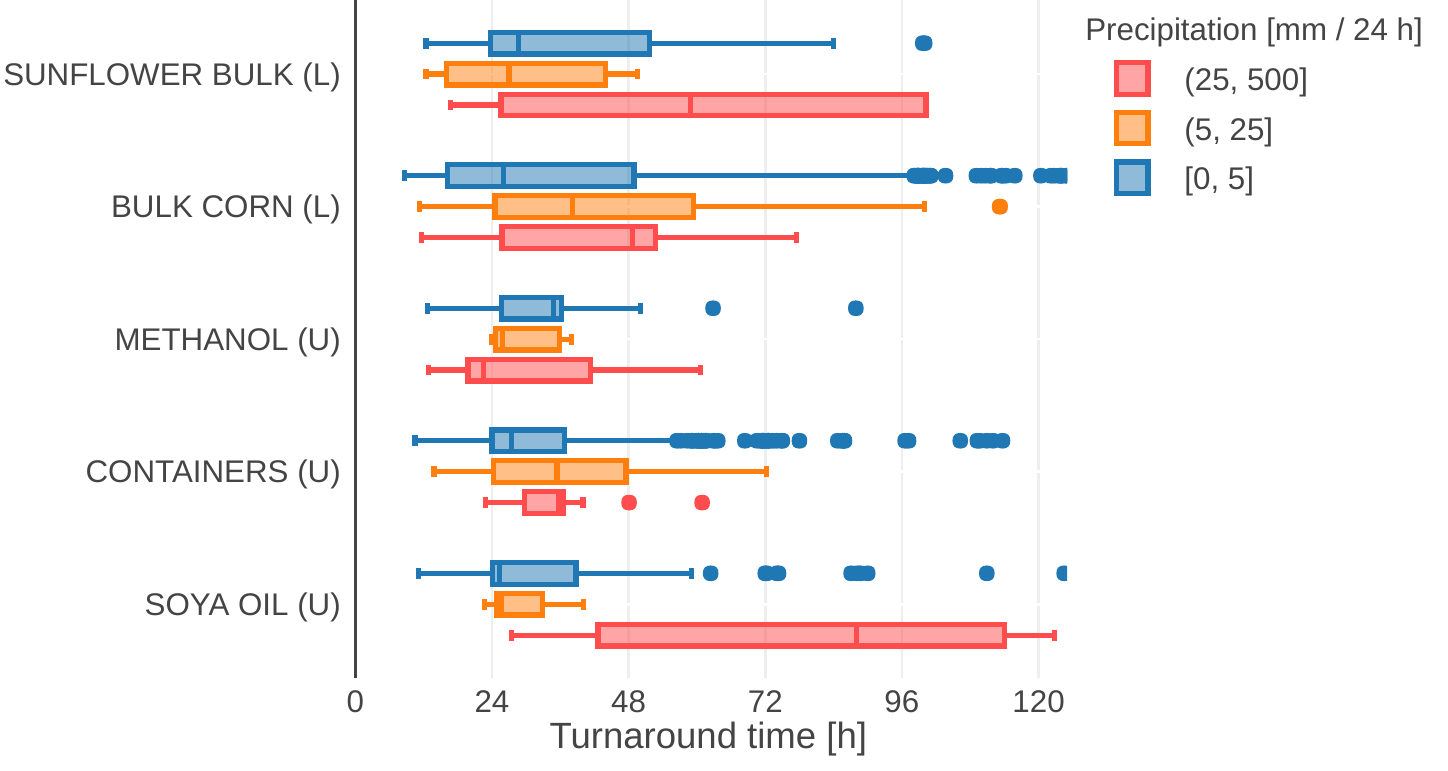}} 
\caption{Influence of different levels of precipitation on turnaround time of different loaded and unloaded cargo types.}
\label{fig:gpmb_weather_turnaround_time}
\end{figure}

We also experimented with weather data, especially wind and precipitation data, that should have influence on the turnaround time. We obtained hourly aggregated weather data and combined it with historical port call data. We did notice that the level of precipitation has the influence, as presented in Figure~\ref{fig:gpmb_weather_turnaround_time}. Certain dry bulk cargo types experience noticeably longer turnaround times, in comparison with containers or liquid cargoes. Nevertheless, when we presented such features to the predictive model, it did not result in an increased accuracy. This might again be specific to the Port of Bordeaux.

\section{Conclusion}

In this paper we present a novel machine learning based system for turnaround time prediction, based on standardized port call data, captured in the ports. We evaluate the proposed system on the real-world example of the Port of Bordeaux, consisting out of 11 years worth of historical data, as well as live operational data. Our results show that port call data can be successfully used to predict turnaround time, with predictions for certain cargo types reaching error rates below 10\%, compared to ground truth data. We also show that our proposed data driven approach significantly outperforms currently used manual approach in the port. We also demonstrate, that port call data offers an insight into port operations, by exploring the collected data to obtain various business metrics, some of them presented in our exploratory data analysis.

\section*{Acknowledgment}

This work was partially supported by the European Commission through the Horizon 2020 research and innovation program under grants 769355 (PIXEL).

\bibliographystyle{IEEEtran}
\bibliography{IEEEabrv,IEEEexample}


\end{document}